\documentclass[10pt]{wlscirep}
\usepackage[utf8]{inputenc}
\usepackage[T1]{fontenc}
\usepackage{lineno}
\usepackage{subcaption}
\usepackage{graphicx}
\usepackage{ctable}
\usepackage{booktabs}

\usepackage{subcaption}
\usepackage{array}
\newcolumntype{L}[1]{>{\raggedright\let\newline\\\arraybackslash\hspace{0pt}}m{#1}}
\newcolumntype{R}[1]{>{\raggedleft\let\newline\\\arraybackslash\hspace{0pt}}m{#1}}
\usepackage{multirow}
\newcommand\Tstrut{\rule{0pt}{2.5ex}}         % = `top' strut

\title{StreetSurfaceVis: a dataset of crowdsourced street-level imagery annotated by road surface type and quality}
\author[1,*]{Alexandra Kapp}
\author[1]{Edith  Hoffmann}
\author[1]{Esther Weigmann}
\author[1]{Helena Mihaljević}
\affil[1]{Hochschule für Technik und Wirtschaft Berlin (HTW Berlin)}

\affil[*]{corresponding author(s): Alexandra Kapp (alexandra.kapp@htw-berlin.de)}

\begin{abstract} %170 words
Road unevenness significantly impacts the safety and comfort of traffic participants, especially vulnerable groups such as cyclists and wheelchair users.
To train models for comprehensive road surface assessments, we introduce \textit{StreetSurfaceVis}, a novel dataset comprising 9,122 street-level images mostly from Germany collected from a crowdsourcing platform and manually annotated by road surface type and quality. 
By crafting a heterogeneous dataset, we aim to 
enable robust models that maintain high accuracy across diverse image sources. 
As the frequency distribution of road surface types and qualities is highly imbalanced, we propose a sampling strategy incorporating various external label prediction resources to ensure sufficient images per class while reducing manual annotation.  
More precisely, we estimate the impact of (1) enriching the image data with OpenStreetMap tags, (2) iterative training and application of a custom surface type classification model, (3) amplifying underrepresented classes through prompt-based classification with GPT-4o and (4) similarity search using image embeddings. Combining these strategies effectively reduces manual annotation workload while ensuring sufficient class representation.
\end{abstract}
\begin{document}

\flushbottom
\maketitle

\thispagestyle{empty}

\section*{Background \& Summary}

%(700 words maximum) 
Road damages have a significant impact on the comfort and safety of all traffic participants, especially for vulnerable road users such as  cyclists~\cite{gadsby_understanding_2022,nyberg_road_1996}, wheelchair users~\cite{pearlman_pedestrian_2013, duvall_development_2013, beale2006mapping} and individuals employing inline skates~\cite{lorimer2015beyond}, cargo bikes~\cite{athanasopoulos2024integrating}, scooters~\cite{rodier2003unsafe}, or strollers. They have also been identified as a major cause of traffic accidents~\cite{kurebwa2019study}.
These issues have sparked a large body of research on methods that apply deep learning models to street-level imagery for road surface condition assessment~\cite{cao_survey_2020, rahman_comparative_2022, kim_review_2022}.
Yet, road damage may not reflect the full range of factors that influence a traffic participant's experience. For example, the smoothness of sett (regular-shaped cobblestone) is rather determined by the flatness of the utilized stones. 
Rateke et al.~\cite{rateke_road_2019} developed a hierarchical vision-based approach, first predicting the surface \textit{type} and then employing specific models for each type to classify \textit{quality}. They utilize the `Road Traversing Knowledge for Quality Classification' (RTK) dataset ~\cite{rafael_toledo_road_2023} comprising 6,264 images captured in a Brazilian city with a low-cost camera setup attached to a moving vehicle that was annotated by surface type and quality. However, the model trained on the RTK dataset does not generalize well to other datasets~\cite{rateke_road_2019}, likely due to the lack of image heterogeneity. 
Similar to the RTK dataset, typical street-level imagery datasets are commonly collected in good weather conditions, using only a single vehicle and camera setup within a limited geographic boundary, e.g.,  KITTI~\cite{doi:10.1177/0278364913491297}, an autonomous driving benchmark dataset from a mid-sized city in Germany, CaRINA~\cite{7795529} a road surface detection dataset from São Carlos in Brazil, or Oxford RoboCar~\cite{maddern20171}, a dataset of 100 repetitions of a consistent route through Oxford, UK, to capture different weather conditions. Cityscapes~\cite{cordts_cityscapes_2016} is a dataset of 25,000 images of street scenes recorded in 50 mostly German cities tailored for autonomous driving applications. Even though it provides more diversity than comparable datasets, perspectives of sidewalks and cycleways are not considered and labels consist of semantic segmentation, while surface type and quality information is not available.

This paper introduces \textit{StreetSurfaceVis}, a new street-level image dataset comprising 9,122 images with a substantial amount for each pertinent surface type and quality,
%labeled with surface type and quality
fostering the training of robust classification models. 
We utilize the crowdsourcing platform \textit{Mapillary}~\cite{noauthor_mapillary_2024} to gather images, as these are contributed by individuals from various regions using different devices, camera angles, and modes of transportation, resulting in a heterogeneous dataset. Surface type and quality are labeled by human experts.
%However, the challenge lies in efficiently obtaining \textit{labels} for type and quality. 
According to data from the crowdsourcing geographic database \textit{OpenStreetMap} (OSM), distributions of surface type and quality are highly skewed in Germany: For instance, asphalt is the predominant road type, accounting for 47\% of tagged road segments, while only about 3\%  are made of sett. Similarly, for 54\% of asphalted roads, the quality is considered `good', while only 1\% is rated `bad'. Even though OSM data is incomplete, we assume the distribution to be a reasonable approximate estimate. Consequently, the difficulty lies in gathering sufficient images for every relevant class without an infeasible manual labeling effort.
%This paper introduces \textit{StreetSurfaceVis}, a new street-level image dataset comprising 9,122 images from the crowdsourcing platform Mapillary, labeled for road surface type and quality by human experts. 
Thus, we present and evaluate different strategies for semi-automated annotation to efficiently amplify underrepresented classes in the dataset. These strategies include (1) pre-filtering using OSM tags, (2) iterative training and application of type classification models, (3) prompt-based image classification with GPT-4 models, and (4) similarity-based search using image embeddings.

\section*{Methods} \label{sec:dataset_construction}

\subsection*{Image base} Our dataset is based on images from Mapillary
%~\cite{noauthor_mapillary_2024}
limited to the geographical bounding box of Germany. Launched in 2013, this crowdsourcing platform provides openly available street-level images. Contributors can use, among others, the Mapillary smartphone app to capture georeferenced image sequences during their trips by, e.g.,  car, bicycle, or on foot. Thus, the dataset encompasses not only images from roadways but also cycleways and footways. 
As of January 2024, Mapillary contains about 170 million images in Germany, including hundreds of thousands for every major city, with over 50\% captured within the last three years.
The geographic coverage varies depending on the contributors within each region. 
Moreover, the dataset shows a wide range of quality influenced by factors like the device used, its positioning (e.g., a visible car dashboard), or the prevailing lighting and weather conditions.
As a result, images may exhibit varying degrees of darkness, sharpness, or blurriness, and may include (or even focus on)  additional objects, such as traffic signs, cars, or trees.
%Hence, some images are dark, blurred, or contain motifs other than street-level scenes, such as traffic signs, cars, or trees. 

\subsection*{Image selection} \label{subsec:imageSelection}
Mapillary images are typically captured during a trip via the Mapillary app (or another recording method) which captures images every few seconds, creating a sequence of images with a shared identifier. 
To increase the dataset's heterogeneity, we limit the number of images for each location and sequence per type and quality class. 
%A sequence pertains to the recording during a single trip where the Mapillary app captures images every few seconds. 
This reduces the number of images taken by the same person on one trip and thus increases spatial diversity, camera specifications, environmental conditions, and photographic perspectives. Specifically, we limit the number of images per geographic unit (XYZ style spherical Mercator tiles on zoom level 14,  which is roughly equivalent to $\sim 1.5$ x $1.5 km^2$ grid cells. We thereby adhere to the same geographic unit as utilized by the Mapillary API for computational feasibility) to 5 and the number of images per sequence to 10.

%Five German cities, varying in region and population size -- Munich, Cologne, Lunenburg, Dresden, and Heilbronn -- have been exclusively selected to comprise the test data, each contributing about 150 images. 
%Thus, the capability of models to classify images from yet unseen cities can be tested.

\subsection*{Labeling scheme}
Our labels for surface \textit{type} and \textit{quality} primarily align with the \textit{OSM} road segment tags \texttt{surface} and \texttt{smoothness}, respectively.  
%OSM is a free and open geographic database maintained by a community of volunteers. 
While \texttt{surface}~\cite{noauthor_osm_2024-1} describes the surface type such as `asphalt', \texttt{smoothness}~\cite{noauthor_osm_2024} reflects the physical usability of a road segment for wheeled vehicles, particularly regarding its regularity or flatness~\cite{noauthor_osm_2024}.
Our labeling scheme includes those classes that are important from a traffic perspective and represent a relevant portion of street types in Germany. This results in the type labels \textit{asphalt}, \textit{concrete}, \textit{paving stones}, \textit{sett}, and \textit{unpaved}
\footnote{More precise options for unpaved include ground, (fine) gravel, grass, compacted, and dirt, but this level of differentiation is not relevant for our context.}, each of which accounts for at least 1\% of the tagged road segments.
For the quality label, we restrict to five of eight proposed levels, ranging from \textit{excellent} (suitable for rollerblades), \textit{good} (racing bikes), \textit{intermediate} (city bikes and wheelchairs), \textit{bad} (normal cars with reduced velocity) to \textit{very bad} (cars with high-clearance). 
The final scheme comprises 18 classes of type and quality combinations, as not all quality labels are suitable for all surface types. See Figure~\ref{fig:exampleImg} for example images and labels and Table~\ref{tab:labels} for descriptions for each class.
%\subsection{Manual annotation}

\subsection*{Manual annotation} After conducting a thorough explorative analysis of Mapillary images, the first three authors developed an annotation guide containing quality level descriptions with example images and underwent self-organized training to manually label surface type and quality. The instructions include labeling the focal road located in the bottom center of the street-level image. In cases where the focus is ambiguous, such as when two parts of the road (e.g., the cycleway and footway) are depicted equally, or when the surface could not be classified due to factors such as snowy roads, blurry images, or non-road images, the image is sorted out.
If the surface quality falls between two categories, annotators are directed to select the lower quality level. Annotators are encouraged to consult each other for a second opinion when uncertain.
For annotation, we use the tool \textit{Labelstudio}~\cite{LabelStudio}, which allows to preset labels from pre-labeling strategies.
%\textit{Labelstudio}~\cite{LabelStudio} was selected as a tool to facilitate the annotation process. It allows the pre-setting of labels, thus, providing the possibility to incorporate results of pre-labeling strategies. 

%\footnote{`not recognizable' is not included as a separate class within the dataset and annotators were encouraged to exclude images when in doubt.}. 
%and only the 119 images with valid labels from all annotators were included to access the level of agreement
 %`not recognizable' is not deemed a class and but instead an exclusion criteria for the dataset, considering this as a class would only reflected differences in the annotators' perception of edge cases on which images can still be considered `recognizable' instead agreement on class ratings.}.

\subsection*{Annotation strategies}

We assume highly uneven class distributions and use the frequency distribution of OSM tags pertaining to surface type and quality as a baseline estimate\footnote{The OSM distribution is likely rather an overestimation of the frequency of underrepresented classes, as main roads with good quality are typically more frequented, and thus presumably have more images.}. According to this data, manual labeling of randomly sampled images would be highly inefficient. For example, as only 0.7\% of road segments are tagged as \textit{asphalt-bad}, we would require manual checking of 1,000 images to  obtain 7 \textit{asphalt-bad} images (cf. Table ~\ref{tab:improvements_1}). Thus, our goal is to employ pre-selection strategies that yield samples for manual labeling where underrepresented classes have a substantially higher frequency than indicated by the OSM baseline. 
We evaluate four strategies: 
(1) enriching the image dataset with OSM tags; (2) iterative training and application of a model classifying surface type; and (3) amplifying underrepresented type-quality classes using GPT-4 prompts and (4) similarity search based on image embeddings. In the following, we describe each strategy and evaluate its impact. 
Our proposed overall approach is depicted in Figure~\ref{fig:ds_construction}.  
Table ~\ref{tab:improvements_1} presents the improvements achieved through the first two strategies in terms of precision*100 %, based on 19,747 images
(marked in green and yellow in Figure~\ref{fig:ds_construction}). 

\subsubsection*{Pre-labeling via OSM tags.}
label{subsec:pre-selection}

OSM ~\cite{noauthor_openstreetmap_2024} contains \texttt{surface} tags for 5249\%  and \texttt{smoothness} tags for 8,6\% of road segments in Germany, as of August 2024. We incorporate this information by spatially intersecting with geolocations of Mapillary images\footnote{For computational feasibility, we refrain from intersecting all Mapillary images and use a sample of tiles where the desired classes occur particularly frequently.} and assigning the \texttt{surface} and \texttt{smoothness} labels of the closest OSM road segment within a maximum distance of two meters.
To eliminate ambiguous street intersections, we cut off 10\% of the start and end of each road segment beforehand. We refer to the labels resulting from this strategy as \textit{OSM pre-labels}.

Using a sample of 100 images for each of the 18 classes according to OSM pre-labels (Batch 1 in Figure ~\ref{fig:ds_construction}), OSM pre-labels and manual annotation agree on 69\% of surface type labels, and of these, the quality is correct for 55\% of the images. %However, the errors are not equally distributed: while 94\% of asphalt images are correctly pre-labeled, only 54\% of paving stones and sett are correct. 
Incorrect type labels mainly result from mixing up adjacent road parts, as OSM sometimes lacks separate geometries for roadways, cycleways, and footways. 
Differences in quality labels are likely due to varying subjective assessments by OSM contributors. Additionally, GPS inaccuracies as well as time differences between image capturing and road segment tagging in OSM are plausible sources of discrepancies, for both type and quality.
Even though many pre-labels are incorrect, this strategy increases class precision substantially, 
%, albeit with varying impact. 
for example, to obtain 7 images from the underrepresented class \textit{asphalt-bad}, around 100 instead of estimated 1,000 images need to be reviewed (cf. Table ~\ref{tab:improvements_1}). (Note that this estimate is based on 19,747 images, out of 5M Mapillary records, that were used as input for the next strategy.)

\subsubsection*{Pseudo-labeling with a type classifier.}
\label{subsec:pseudolab}

To further increase pre-selection precision, we iteratively train a classification model to predict the type and use its predictions as pseudo-labels.
Since type is easier to classify than quality, a smaller amount of data should be sufficient to obtain valuable pseudo-labels. 
More precisely, we fine-tune EfficientNetV2-S~\cite{tan2021efficientnetv2}, pre-trained on ImageNet, on the first annotated batch to predict the type.
The model is then applied to the next batch of images. All images where the prediction matches the OSM pre-label are selected for manual annotation; this combination of labels achieves an average precision of 95\% for the surface type, with the lowest precision of 89\% for paving stones over all batches. To reduce bias towards easy-to-classify examples, a random sample of 10\% from the excluded images is manually annotated. 
In the next iteration, the training set is extended with the manually annotated images from the previous round. In subsequent iterations, the batch composition is adjusted towards underrepresented classes, aiming for 300-400 images per class. 

We ceased this procedure after including a substantial amount of images according to OSM pre-labels for every class, resulting in a dataset of 7,033 images. This required applying the type prediction models to 19,747 images filtered based on OSM pre-labels and  manually annotating 8,175 images. 

This procedure provides significant improvements for certain classes, for example, \textit{paving stones-excellent} achieves a precision of 30\% (cf. Table ~\ref{tab:improvements_1}) with a total of 342 images at this point.  
However, the precision remains very low for some classes, e.g., less than 4\%  for \textit{paving stones-bad} with only 30 images collected at this point, while an excessive amount of images from overrepresented classes remains within the sample for manual annotation, %like \textit{asphalt-good} images 
with e.g. 1,334 images labeled as  \textit{asphalt-good} at this point. 
 Thus, continuing this procedure to sufficiently represent all classes would be infeasible.

\subsubsection*{Prompt-based image classification and similarity search.}

We evaluate two approaches to efficiently enlarge classes that remain underrepresented. Our first approach uses \textit{prompt-based image classification} with OpenAI's GPT-4V~\cite{noauthor_gptv_system_card_2024} and GPT-4o~\cite{noauthor_hello_2024} models, which can generate textual output from image-text input. 
Previous studies have demonstrated the potential of GPT-4V for automated image labeling in various application domains, including 
%bio-medicine  ~\cite{liu_holistic_2023,hou_gpt-4v_2024}, 
street intersections \cite{hwang_is_2024} and traffic scenery \cite{wen_road_2023}. 
GPT-4o, released in May 2024, is expected to be similarly effective at half the cost (at the time of release).
Despite this and possible future cost reductions, inference with any of these models implies ongoing monetary expenditure. 
As an alternative, we explore  \textit{similarity search} using image embeddings \cite{coleman_similarity_2022} from OpenAI's CLIP~\cite{radford_learning_2021}, DINOv2~\cite{oquab_dinov2_2023}, and our fine-tuned EfficientNet-based type classifier. Specifically, annotated images from the class of interest are used as the query, and all images with a cosine similarity score above a certain threshold are pre-labeled as class members. 

For both approaches, we restrict the search space using the previously described strategy combining OSM tags and type classification, as these have proven to be efficient low-cost strategies. Due to cost and time constraints, we limit the experiments to three underrepresented classes. 
Using a validation dataset of randomly sampled manually annotated images (50 for the classes \textit{asphalt-bad} and \textit{paving stones - intermediate} and 30 for \textit{paving stones - bad}), we systematically 
evaluate base models and hyperparameters of both approaches\footnote{Note, that for \textit{paving stones - bad} only 30 images were available after applying the first two strategies.}. 
Note that in both cases, a higher precision implies a lower estimated effort in human post-annotation, while a higher recall correlates with a larger class increase and, in the prompting scenario, with a lower monetary cost.

Prompting configurations include two different image cropping styles, varying levels of detail in class definitions, zero-shot versus one-shot prediction, and processing a batch versus one image per prompt. While cropping shows only minimal effect, one-shot outperforms the zero-shot setting, albeit with a larger monetary cost, resulting in a similar price per hit. GPT-4o achieves notably better results than GPT-4V for all classes. The best configuration in terms of F1 score and cost-effectiveness consists of GPT-4o, lower half-center cropping, shortened definitions, one-shot prediction, and one request per prompt. The final prompt for the example of surface type asphalt is provided in Table~\ref{tab:prompt}.
%\footnote{See final prompt here: \url{https://github.com/SurfaceAI/dataset_creation/tree/main/src/scripts/gpt_experiments/example_prompt.md}}. 
%Estimated precision (recall) values based on the described validation set ranged from 0.50 to 1 (0.27 to 0.70); and the estimated price per hit ranged from \$0.06 to \$0.94 depending on the class.

To decide on the similarity threshold, we compute the optimal ROC cut-off value for each embedding model and class. 
All three embeddings give similar and reasonable results for the two larger classes with optimal cut-off values around 0.85 for CLIP and 0.6 for the other two models. The type-classifier slightly outperforms DINOv2 and CLIP on average and is therefore selected for the following experiment.
The smallest class \textit{paving stones - bad} remains difficult for all three models, e.g., our embedding returns nearly all input images (of type paving stones) given the optimal threshold.
%, i.e., the score that maximizes the difference between the true positive and the false positive rate. 
%The resulting precision and recall values ranged from X to Y, respectively.

Both the prompting and similarity search approach are then applied in their optimal configurations to a dataset compiled as follows: around 20M Mapillary images are used as a base,  filtered according to the number of images per tile and sequence, as described above, and pre-labeled using OSM tags and type pseudo-labels. From the remaining data, up to 1,000 images per pre-label are drawn at random. Note that for the smallest class \textit{paving stones-bad} only 210 images are available after applying these two strategies. 

As shown in Table \ref{tab:improvements_2}, both approaches yield a substantial improvement. However, prompting with GPT-4o achieves substantially higher precision than the similarity search, with values between 40\% and 65\% in comparison to 10\% to 31\%\footnote{Note that for consistency with OSM frequency distribution values (cf. Table~\ref{tab:improvements_1}) we report precision and recall in percentage, as this would otherwise require up to 4 decimals}.  Note, that the recall of 100\% for \textit{paving stones-bad} in the similarity search is due to almost all images being classified as true, which also results in a precision similar to the baseline. Thus, this method is not suitable for this class.
%Except for \textit{paving stones-bad}, the recall is also higher. 

While the described strategy has shown to have a notable impact (cf. Table~\ref{tab:improvements_2}), it substantially limits the search space %(for both GPT-4o and the similarity search) 
due to the sparsity of OSM tags. 
%on as only 16\% of OSM road segments are tagged with regard to \texttt{smoothness}; as already seen in the small search space for \textit{paving stones-bad}. 
Moreover, the reliance on OSM tags yields a potential selection bias as the distribution can be assumed to depend on factors such as urbanity and OSM community. 
To estimate the efficacy of GPT-4o prompt-based classification without OSM-based filtering, we conduct an additional experiment on a search space obtained only by pre-selecting the type according to the type classification model on a random sample of 20,000 Mapillary images. From the resulting 15,100 images classified as asphalt, we prompt GPT-4o with a random sample of 2,000 images, as well as all 712 images pseudo-labeled as paving stones.
The results, depicted in the last column of Table~\ref{tab:improvements_2}, show a reduced precision for each class, and thus imply a higher manual labeling effort, with the largest decrease for \textit{paving stones - bad} from 64.7\% to 18.2\%. 
Nevertheless, there remains a relevant increase to the OSM pre-label baseline, providing a viable method if the OSM tag pre-labeled search space is exhausted.
However, costs substantially rise: with about \$0.01 per GPT-4o prompt, obtaining a correctly labeled image without pre-labeling via OSM costs \$0.12 to \$3.45, depending on the class, compared to \$0.07 and \$0.18 in the previous experiment.
%and a substantial reduction of hits (below 1\% for both \textit{bad} classes) which result in respective cost increases, with costs per hit ranging from \$0.12 to \$3.45. 
%Note that both datasets do not intersect with the thus far labeled data used for the training of the classification model and as query images for CLIP.
%For each of the two vision-model-based approaches, all images predicted to fall into each of the classes of interest were annotated manually. 
Overall, we achieved a substantial increase of instances in underrepresented classes, as shown in  Table~\ref{tab:labels}\footnote{Note that following the evaluation we utilized the proposed methods for expanding the underrepresented classes \textit{concrete-bad} and \textit{sett-good} as well.}, with a major reduction of manual labeling effort. However note, that there remain class sizes far below the target class size of 300-400 images, likely due to their low occurrence on German roads, which aligns with the estimate based on OSM.

Generally, performance increases of GPT-4o and similar models are to be expected in the future, further amplifying the viability of this approach. To reduce dependency on OpenAI and monetary cost, future work should evaluate open-source alternatives. 
Similarity search is more efficient in computational and monetary terms, therefore currently remaining a viable alternative despite inferior results. The results varied between classes and showed the worst performance for \textit{paving stones - bad}, where there was no improvement to the baseline. Further experiments, especially with more images for the smallest class, could be explored, as well as advancements such as incorporating clustering strategies \cite{vo_automatic_2024}.

%We carefully filtered images based on metadata information to enhance diversity. However, especially the limitation of OSM \texttt{smoothness} tags causes sampling of multiple images from identical roads for underrepresented classes.
%Despite our efforts to enhance underrepresented classes, the total counts of certain classes remain low. This might indicate the minor relevance of these classes for the scope of Germany.

%We envision extensions of the dataset to include road-type information as well as unsuitable images, such as blurred ones or those without a road in focus. Including these will enable training models to automatically check image suitability, thus improving classification robustness. 

\section*{Data Records}

%The Data Records section should be used to explain each data record associated with this work, including the repository where this information is stored, and to provide an overview of the data files and their formats. Each external data record should be cited numerically in the text of this section, for example \cite{Hao:gidmaps:2014}, and included in the main reference list as described below. A data citation should also be placed in the subsection of the Methods containing the data-collection or analytical procedure(s) used to derive the corresponding record. Providing a direct link to the dataset may also be helpful to readers (\hyperlink{https://doi.org/10.6084/m9.figshare.853801}{https://doi.org/10.6084/m9.figshare.853801}).

%Tables should be used to support the data records, and should clearly indicate the samples and subjects (study inputs), their provenance, and the experimental manipulations performed on each (please see 'Tables' below). They should also specify the data output resulting from each data-collection or analytical step, should these form part of the archived record.

StreetSurfaceVis is an image dataset containing 9,122 street-level images within Germany's bounding box with labels on road surface type and quality; find the number of instances per class in Table \ref{tab:labels}. A csv file contains all the image metadata, and four folders contain the image files. Based on the image width, all images are available in four different sizes: 256px, 1024px, 2048px, and the original size.
Folders containing the images are named according to the respective image size. Image files are named based on the \texttt{mapillary\_image\_id}.
This repository ({\url{https://doi.org/10.5281/zenodo.11449977}) provides the dataset, a description, the labeling guide, and a datasheet documenting the dataset~\cite{kapp_2024_11449977}.

\section*{Technical Validation}

\subsection*{Inter-rater reliability}
To evaluate inter-rater reliability, 180 images (10 images per class according to OSM pre-labels) were independently rated by all three annotators.
Twelve images were marked for revision by at least one annotator, and another 49 were discarded by at least one annotator for the above reasons, all of which were excluded from the calculation of inter-rater reliability.
Krippendorff's $\alpha$~\cite{krippendorff_content_2019} for surface type is calculated at 0.96, indicating a high level of agreement. 
%\footnote{Krippendorff's $\alpha$=0.82 if `not recognizable' was considered as a class}. 
Surface quality, treated as an ordinal scale variable, achieves a Krippendorff's $\alpha$ of 0.74.
While this is generally deemed an acceptable level of agreement, it reflects the fluid class transitions of quality in contrast to type. 

%- annotation guide development
%- inter-rater reliability
%- Modelle type / quality

\subsection*{Type and quality model performance}

To assess the validity of our dataset, we train an EfficientNetV2-S-based model to predict surface types. 
We split the final dataset into a training set of 8,346 images and a test set of 776 images from five cities geographically distinct from the training data. Note that we do not enhance underrepresented classes in the test data, aiming to reflect real-life distributions. 
 traapplyvan 80:20lidation split of 80:20 and conduct five runs with different seeds, especially influencing the train-validation split, and report the averaged results.
 
We use the validation dataset solely to identify the optimal number of epochs, without tuning other hyperparameters. 
An accuracy (loss) of 0.96 (0.13) is achieved for the training data, 0.94 (0.19) for validation, and 0.91 for test data, respectively. Table~\ref{tab:modelEval} presents the recall, precision, and F1 scores of the test data for each surface type. 
All F1 scores for the test data are equal to or exceed 0.9, except for the `concrete' surface type. This demonstrates a strong generalization of our training data to Mapillary images from previously unseen (German) cities. The low F1 score of 0.35 for concrete can be attributed to its visual similarity to asphalt and its rare occurrence in the dataset. Consequently, a small portion of the large asphalt class is misclassified as concrete. Given the limited number of concrete images, this results in a low precision for the concrete class. Depending on the application, e.g., surface classification for routing purposes, distinguishing between concrete and asphalt may not be needed, and the two classes could be merged into one.

In a similar setup, we train five regression models, one for each type, to predict surface quality, using mean squared error as the loss function. For evaluation, we assume a correctly classified type, i.e., the quality prediction is independent of the type model.
Deviations from the true value are normally distributed and centered around 0, with an overall Spearman correlation coefficient of 0.72 and type-specific coefficients between 0.42 and 0.65 (see Table~\ref{tab:modelEval}), thus moderate to strong correlations. When converting numeric predictions into quality categories, the models  achieve an overall accuracy of 0.63, with type-specific accuracies ranging from 0.61 to 0.71. To account for fluidity in quality annotation, we also compute the 1-off accuracy which considers neighboring classes as correct classifications. All 1-off accuracies are (almost) 1.0, showing that all model predictions are at most one class off, demonstrating a high level of precision.
%Figure~\ref{fig:deviation} shows the distribution of deviations between true and predicted values for each type. xx\% of all predictions do not deviate more than $\pm 0.5$ and only xx\% more than $\pm 1$, which is equivalent to a mismatch with the neighboring class. The Spearman correlation coefficients for surface qualities for each type range between xx and xx (see Table~\ref{tab:modelEval}).
%We evaluate the regression based on the root mean squared error (RMSE). Note, that for types with less quality classes (e.g., sett only has three classes) the RMSE will naturally be smaller.
%An RMSE of xx\% is achieved for the training data, xx\% for validation and xx\% for test data, respectively. 

\subsection*{Cross-dataset generalization testing}
To demonstrate the ability of our training dataset to train models that generalize to other data sources, we train the type and quality models on our dataset to predict the RTK~\cite{rafael_toledo_road_2023} dataset and vice versa. RTK contains low-resolution street-level images captured with one moving vehicle in a Brazilian town. Images are labeled according to type (asphalt, paved, and unpaved) and quality  (good, regular and bad), resulting in the following instances:
asphalt-good: 1,978, asphalt-regular: 839, asphalt-bad: 464, paved-good: 1,179, paved-regular: 324, paved-bad: 124, unpaved-bad: 593, and unpaved-regular: 796.

We merge our asphalt and concrete images to a single class, matching the RTK  `asphalt' class, and, similarly, paving stones and sett are merged into `paved'. As we formulate the quality prediction as a regression problem,  we can utilize the Spearman correlation coefficient to compare true and predicted values and thus do not need to match labels.
Since our dataset is larger, we down-sample our dataset to match the RTK image count of 6,297 when utilized for training while maintaining our class distribution. Again, we train each model five times with different seeds and report averaged results. As there are different class sizes between models, we report the average (unweighted) F1 score as an overall metric. We determine significance according to a two-sided Mann-Whitney U test (nonparametric alternative of the t-test) with a significance level of 0.05. 

As shown in Table~\ref{tab:crossDatasetRes}, the model trained on our dataset achieves a significantly higher average F1 score of 0.81 compared to 0.56 of the model trained on RTK. While the RTK model slightly outperforms our dataset on the recall of paved roads, it performs poorly on the detection of unpaved roads.
Note that we trained the models in a vanilla setting, without applying additional techniques such as blurring augmentation, which is expected to enhance the performance of the model trained on StreetSurfaceVis due to the higher resolution of the images.

%We average the Shannon entropy for each image over all five runs to assess the robustness of model predictions. A robust model would obtain similar model predictions independent of varying random seeds~\cite{madhyastha2019modelstabilityfunctionrandom}, thus a low entropy, i.e., low diversity of predictions per image. 

To compare surface quality predictions, we consider the Spearman correlation coefficient between true types and model predictions: overall model predictions based on our dataset achieve a coefficient of 0.52, while it is  0.16 vice versa, indicating that the model based on our dataset captures quality differences of the RTK dataset, while not the other way round. See Table~\ref{tab:crossDatasetRes} for type-wise correlation coefficients.
Note, that quality prediction works best for asphalt roads which is not surprising as the quality definitions for paved and unpaved do not entirely align between both datasets.

%This section presents any experiments or analyses that are needed to support the technical quality of the dataset. This section may be supported by figures and tables, as needed. This is a required section; authors must present information justifying the reliability of their data.

\section*{Usage Notes}

%The Usage Notes should contain brief instructions to assist other researchers with reuse of the data. This may include discussion of software packages that are suitable for analysing the assay data files, suggested downstream processing steps (e.g. normalization, etc.), or tips for integrating or comparing the data records with other datasets. Authors are encouraged to provide code, programs or data-processing workflows if they may help others understand or use the data. Please see our code availability policy for advice on supplying custom code alongside Data Descriptor manuscripts.

%For studies involving privacy or safety controls on public access to the data, this section should describe in detail these controls, including how authors can apply to access the data, what criteria will be used to determine who may access the data, and any limitations on data use. 

\subsection*{Train-test split}

For modeling, we recommend using a train-test split where the test data includes geospatially distinct areas, thereby ensuring the model's ability to generalize to unseen regions is tested. We propose urban areas of five cities varying in population size and from different regions in Germany for testing, comprising 776 images tagged accordingly.

\subsection*{Cropping}

As the focal road located in the bottom center of the street-level image is labeled, we recommend to crop images to their lower and middle half prior using for classification tasks.

This is an exemplary code for recommended image preprocessing in Python:

\begin{verbatim}
    from PIL import Image
    img = Image.open(image_path)
    width, height = img.size
    img_cropped = img.crop((0.25 * width, 0.5 * height, 0.75 * width, height))
\end{verbatim}

\subsection*{License}
Images are obtained from Mapillary, a crowd-sourcing plattform for street-level imagery. More metadata about each image can be obtained via the Mapillary API . User-generated images are shared by Mapillary under the CC-BY-SA License.
For each image, the dataset contains the \texttt{mapillary\_image\_id} and \texttt{user\_name}. 
User information can be accessed on the Mapillary website by \url{https://www.mapillary.com/app/user/<USER_NAME>}
and image information by \url{https://www.mapillary.com/app/?focus=photo&pKey=<MAPILLARY_IMAGE_ID>}.
To use the provided images, it is required to adhere to the terms of use of Mapillary\footnote{\url{https://www.mapillary.com/terms?locale=de_DE}}.

\section*{Code availability}

The code for image selection, download and preparation for manual annotation is provided in this repository: \url{https://github.com/SurfaceAI/dataset_creation}.

\bibliography{references}

%\noindent LaTeX formats citations and references automatically using the bibliography records in your .bib file, which you can edit via the project menu. Use the cite command for an inline citation, e.g. \cite{Kaufman2020, Figueredo:2009dg, Babichev2002, behringer2014manipulating}. For data citations of datasets uploaded to e.g. \emph{figshare}, please use the \verb|howpublished| option in the bib entry to specify the platform and the link, as in the \verb|Hao:gidmaps:2014| example in the sample bibliography file. For journal articles, DOIs should be included for works in press that do not yet have volume or page numbers. For other journal articles, DOIs should be included uniformly for all articles or not at all. We recommend that you encode all DOIs in your bibtex database as full URLs, e.g. https://doi.org/10.1007/s12110-009-9068-2.

%\section*{Acknowledgements} (not compulsory)

%Acknowledgements should be brief, and should not include thanks to anonymous referees and editors, or effusive comments. Grant or contribution numbers may be acknowledged.

\section*{Author contributions statement}

A.K. conceived image sampling strategies and mainly wrote the manuscript.
A.K., E.H. and E.W. annotated data.
E.H. and E.W. conducted experiments.
H.M. supervised and advised on all parts. 
All authors reviewed the manuscript. 

%Must include all authors, identified by initials, for example:
%A.A. conceived the experiment(s), A.A. and B.A. conducted the experiment(s), C.A. and D.A. analysed the results. All authors reviewed the manuscript. 

\section*{Competing interests} %(mandatory statement)

The authors declare no competing interests.
%The corresponding author is responsible for providing a \href{https://www.nature.com/sdata/policies/editorial-and-publishing-policies#competing}{competing interests statement} on behalf of all authors of the paper. This statement must be included in the submitted article file.

\section*{Figures \& Tables}

\begin{table}
\caption{Labeling scheme: description for each class.}
\small
\begin{tabular}{llp{13.5cm}}
\toprule
\textbf{Type} & \textbf{Quality} & \textbf{Description} \\
\midrule
\textbf{Asphalt} &excellent & As good as new asphalt, on which a skateboard or rollerblades will have no problem. \\
\hline
&good & Asphalt showing the first signs of wear, such as narrow, smaller than 1.5 cm cracks, or wider cracks filled up with tar, shallow dents in which rainwater may collect, which may cause trouble for rollerblades but not for racing bikes. \\
\hline
&interm. & Asphalt roads that show signs of maintenance, such as patches of repaired surface, wider cracks larger than 2cm. Asphalt sidewalks may contain potholes, but these are small, shallow (<3cm deep) and can be easily avoided. Asphalt driving lanes show damage due to subsidence (depressions of a scale >50 cm) or heavy traffic (shallow ruts in asphalt caused by trucks in summer). This means that the road can be used by normal city bikes, wheelchairs and sports cars, but not by a racing bike. \\
\hline
&bad & Damaged asphalt roads that show clear signs of maintenance: This might include potholes, some of them quite deep, which might decrease the average speed of cars. However, it isn’t so rough that ground clearance becomes a problem. Meaning that the street causes trouble to normal city bikes but not a trekking bike and a car. \\
\hline
\textbf{Concrete}&excellent & As good as new concrete, on which a skateboard or rollerblades will have no problem. \\
\hline
&good & Concrete road showing the first signs of wear, such as narrow, smaller than 1.5 cm cracks, or wider cracks filled up with tar, shallow dents in which rainwater may collect, which may cause trouble for rollerblades but not for racing bikes. \\
\hline
&interm. & Concrete roads that show signs of maintenance, such as patches of repaired surface, wider cracks larger than 2cm. Concrete sidewalks may contain potholes, but these are small, shallow (<3cm deep) and can be easily avoided. Concrete driving lanes show damage due to subsidence (depressions of a scale >50 cm) or heavy traffic (shallow ruts in concrete caused by trucks in summer). This means that the road can be used by normal city bikes, wheelchairs and sports cars, but not by a racing bike. \\
\hline
&bad & Heavily damaged concrete roads that badly need maintenance: many potholes, some of them quite deep. The average speed of cars is less than 50\% of what it would be on a smooth road. However, it isn’t so rough that ground clearance becomes a problem. Meaning that the street causes trouble to normal city bikes but not a trekking bike and a car. \\
\hline
\textbf{Paving st.}&excellent & Newly installed and regularly laid paving stones that show no signs of wear. Gaps may be visible, but they are small and uniform and do not significantly affect the driving experience. \\
\hline
&good & Paving stones showing first signs of wear or newly installed stones with visible but uniform gaps between them. While still suitable for most activities, these surfaces may pose minor challenges for rollerblades and skateboards but remain navigable for racing bikes. \\
\hline
&interm. & Characterized by paving stones exhibiting multiple signs of wear, such as shifted heights, potholes, or cracks. This grade allows for the comfortable passage of normal city bikes and standard vehicles but may prove challenging for racing bikes. \\
\hline
&bad & Heavily uneven or damaged paving stones in dire need of maintenance, featuring significant height disparities and numerous deep potholes. While ground clearance remains sufficient for most vehicles, the surface severely impedes travel, particularly for standard city bikes. \\
\hline
\textbf{Sett}&good & The best sett roads with flattened stones and gaps that are at most small or filled up. The surface might cause trouble for rollerblades but not for racing bikes. \\
\hline
&interm. & The surfaces of the sett stones are not completely flat, or there may be slightly larger gaps between the stones, this causes problems for racing bikes and slows down city bikes and cars. \\
\hline
&bad & Sett stones with large and possibly uneven gaps or uneven stone surfaces or damaged stones, resulting in an overall bumpy surface: This results in a highly uncomfortable driving experience for city bikes. The average speed of cars is less than 50\% of what it would be on a smooth road. However, it isn’t so rough that ground clearance becomes a problem. \\
\hline
\textbf{Unpaved}&interm. & The best unpaved roads that have a compacted surface. This grade allows for the passage of normal city bikes and standard vehicles but may prove challenging for racing bikes. \\
\hline
&bad & Unpaved roads that do not have a smooth and compacted surface but ones that can still be used with a car or trekking bike but not with a city bike. This category also includes hiking paths, which are too narrow for cars. \\
\hline
&very bad & Unpaved roads with potholes, ruts or generally a highly uneven surface not safely passable with a regular passenger car but still passable with an average SUV with higher ground clearance. This category also includes highly uneven hiking paths, which are too narrow for cars. \\
\bottomrule
\end{tabular}
\end{table}

\begin{figure}[ht]
  \centering
    % Subfigure 1
  \begin{subfigure}{0.15\textwidth}
    \centering
    \includegraphics[width=\textwidth]{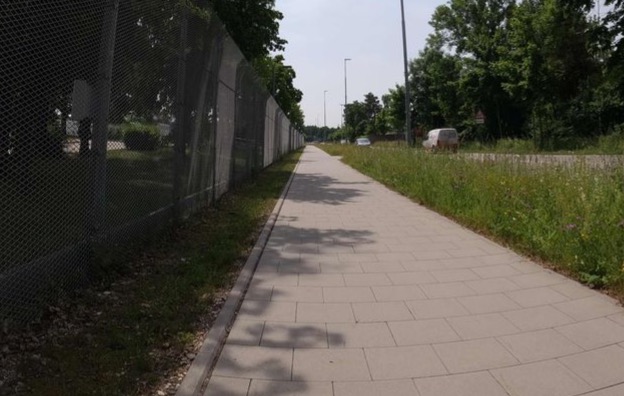}
    \caption{\footnotesize{paving st. - excel.} \newline \tiny{strubbl|3044550732526999}}
    % \includegraphics[width=\textwidth]{img/224717863568050.jpg}
    % \caption{\footnotesize{paving st. - excel.} \newline \tiny{VIA\_KK|224717863568050}}
    \end{subfigure}
    % Subfigure 2
  \begin{subfigure}{0.15\textwidth}
    \centering
    \includegraphics[width=\textwidth]{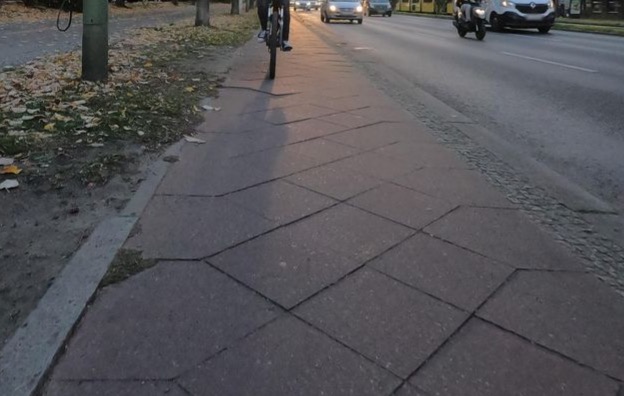}
    \caption{\footnotesize{paving st. - inter.} \newline \tiny{carlheinz|789444188806153}}
    % \includegraphics[width=\textwidth]{img/133978728960378_2.jpg}
    % \caption{pav. stones - inter. \newline \tiny{carlheinz|133978728960378}}
  \end{subfigure}
   % Subfigure 3
  \begin{subfigure}{0.15\textwidth}
    \centering
    \includegraphics[width=\textwidth]{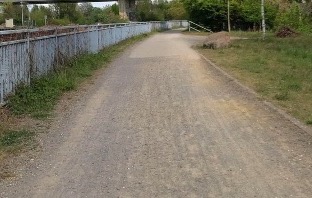}
    \caption{\footnotesize{unpaved - inter.} \newline \tiny{macsico|148771344320493}}
  \end{subfigure}
    % Subfigure 4
  \begin{subfigure}{0.15\textwidth}
    \centering
    \includegraphics[width=\textwidth]{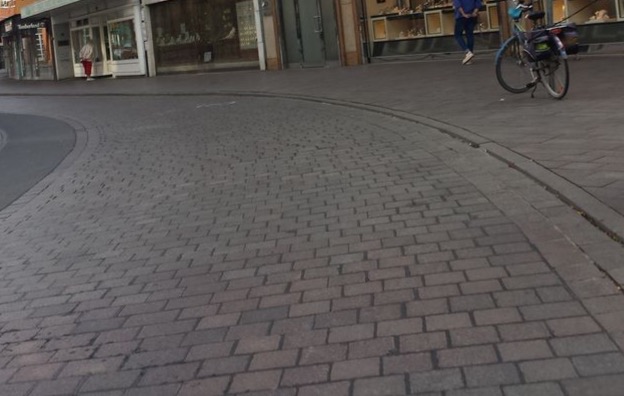}
    \caption{\footnotesize{sett - good} \newline  \tiny{hubert87|812903276289693}}
    % \includegraphics[width=\textwidth]{img/668845388041634.jpg}
    % \caption{sett - good \newline  \tiny{changchun1|668845388041634}}
    \end{subfigure}
    % Subfigure 5
  \begin{subfigure}{0.15\textwidth}
    \centering
    \includegraphics[width=\textwidth]{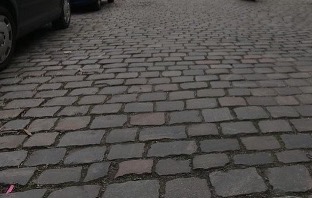}
    \caption{\footnotesize{sett - bad} \newline  \tiny{carlheinz|129178263353193}}
  \end{subfigure}
    % Subfigure 6
  \begin{subfigure}{0.15\textwidth}
    \centering
    \includegraphics[width=\textwidth]{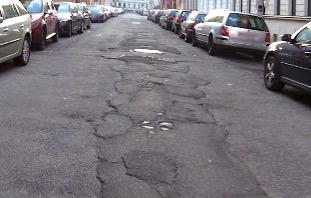}
    \caption{\footnotesize{asphalt - bad} \newline \tiny{zoegglmeyr|293245499014573}}
  \end{subfigure}
  \caption{Example images of different surface types and qualities, with Mapillary contributor names and image IDs.}
  \label{fig:exampleImg}
\end{figure}

\begin{table}[ht]
\centering
\caption{\textit{Precision*100} of (1) OSM pre-label strategy alone and (2) combination with type pseudo-label strategy vs. OSM frequency distribution as a baseline. Note that the numbers refer to different supports, as we utilized 19,747 images for the computation of OSM pre-label precision and 8,175 images for the combination with type pseudo-label.}
\label{tab:improvements_1}
\small
%\begin{tabular}{lR{10mm}|R{12mm}R{23mm}}
\begin{tabular}{l R{2.5cm} R{2.5cm} R{3cm}}
\toprule
%class & OSM \newline distr. & OSM \newline pre-label & OSM  pre-label + \newline type  pseudo-label \\
type-quality class & OSM frequency \newline distribution & OSM pre-label \newline precision & OSM pre-label \newline + type pseudo-label \newline precision \\
\midrule
asphalt-excellent & 20.31 & 27.83 & 43.46 \\
asphalt-good & 29.29 & 28.54 & 50.17 \\
asphalt-intermediate & 4.05 & 14.73 & 20.31 \\
asphalt-bad & 0.69 & 5.64 & 8.80 \\
concrete-excellent & 0.36 & 4.23 & 28.48 \\
concrete-good & 0.94 & 23.32 & 55.03 \\
concrete-intermediate & 0.78 & 22.34 & 44.58 \\
concrete-bad & 0.23 & 9.12 & 20.51 \\
paving stones-excellent & 2.98 & 10.87 & 30.27 \\
paving stones-good & 10.62 & 15.35 & 46.36 \\
paving stones-intermediate & 1.94 & 3.05 & 9.95 \\
paving stones-bad & 0.22 & 1.00 & 3.83 \\
sett-good & 0.56 & 2.19 & 12.92 \\
sett-intermediate & 1.97 & 19.37 & 47.80 \\
sett-bad & 1.26 & 11.92 & 46.65 \\
unpaved-intermediate & 5.75 & 35.09 & 60.87 \\
unpaved-bad & 7.67 & 28.61 & 44.68 \\
unpaved-very bad & 7.38 & 18.70 & 26.89 \\
\bottomrule
\end{tabular}
\end{table}

\begin{figure*}[ht]
    \centering
    \includegraphics[width=0.9\textwidth]{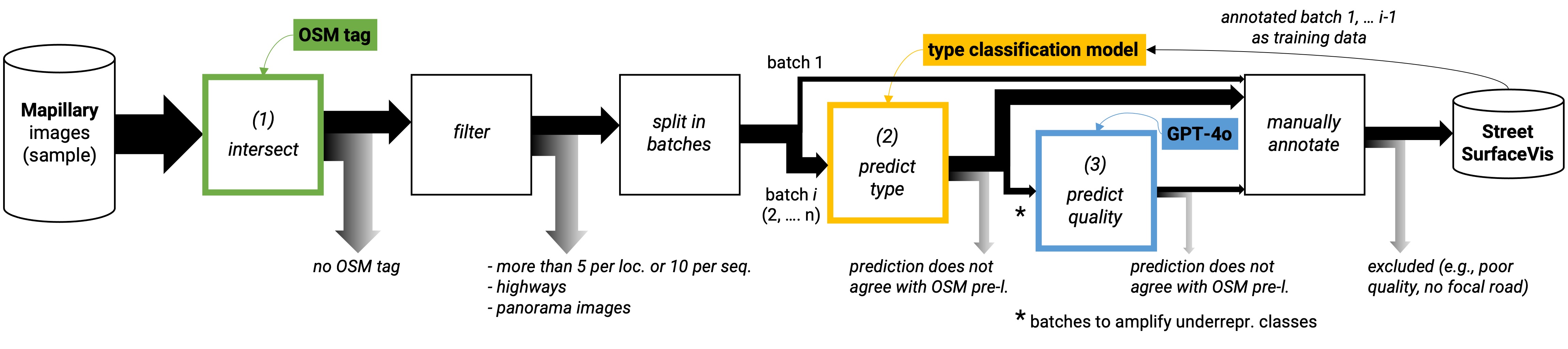}
    %\caption{Process of training dataset creation. Images are manually annotated in an iterative approach such that an initial model prediction can be used for image preselection. Percentages indicate how many images remain from the previous step.}
    \caption{Proposed strategy for selecting, pre-labeling, and annotating the dataset. }
    \label{fig:ds_construction}
\end{figure*}

\begin{table}[t]
%\caption{Evaluation of semi-automated annotation strategies combining OSM tags (OSMT), type classification model (TCM), GPT-4o query, and similarity search using image-embeddings (SimS) on selected underrepresented classes, in terms of \textit{precision*100} | \textit{percentage of hits within search space}. 
%`Hits' refers to the percentage of images of this class within the search space, i.e., how many GPT queries are required to obtain a desired image. 
%OSM prevalence (OSM P) provided as baseline. 
%}
\caption{Evaluation of prompt-based classification (GPT-4o) and similarity search (SimS) for three underrepresented pre-label classes, measured in \textit{precision*100} (\textit{recall*100}), compared to the baseline using OSM pre-labels and type pseudo-labels. The last column shows results for GPT-4o with type pseudo-labels only, excluding OSM pre-labels. Recall is not reported for this configuration and the baseline due to the impracticality of additional manual labeling.}
\label{tab:improvements_2}
\small
\begin{tabular}{L{3.4cm}|R{1.5cm}R{1.5cm}|R{2.2cm}|R{2.2cm}|R{1.5cm}R{2.2cm}}
%\begin{tabular}{lrrrr}
\toprule
%class &  OSM pre-label+\newline type pseudo-label & OSM pre-label+\newline type pseudo-label+GPT-4o & OSM pre-label+\newline type pseudo-label+SimS \\
%\multirow{2}{*}{type-quality class} &  \multicolumn{4} {c|} {search space 1} &  \multicolumn{2} {c} {search space 2} \\
\multirow{4}{*}{type-quality class} &  \multicolumn{4} {c|} {OSM pre-label} & \\
 &  \multicolumn{4} {c|} {+type pseudo-label} &  \multicolumn{2} {c} { type pseudo-label} \\
 & \textit{support} & baseline & + GPT-4o &  + SimS &  \textit{support} & + GPT-4o\\
%class &  OSM pre-l. + \newline type pseudo-l. & OSM pre-l. + \newline type pseudo-l. + GPT-4o & type pseudo-l. + GPT-4o & OSM pre-l. +\newline  type pseudo-l. + \newline SimS \\

\midrule
asphalt-bad &  \textit{1,000} & 10.06 & 39.46 (81.11) & 20.38 \enspace (47.78) &  \textit{2,000} & 21.21  \\
paving stones-bad &  \textit{210} & 10.00 & 64.71 (52.38) & 10.29 (100.00) &  \textit{712} & 18.18 \\
paving stones-intermediate &  \textit{1,000} & 22.75 & 40.29  (83.03) & 30.74 \enspace (47.88) &  \textit{712} & 30.86 \\
\bottomrule
\end{tabular}
\end{table}

\begin{table}[ht]
\caption{Final dataset size by type-quality class. Numbers in parentheses indicate the increase in image counts for underrepresented classes through prompt-based image classification and similarity search.}
\centering
\label{tab:labels}
\small
\begin{tabular}{lrrrrr}
\toprule
& excellent & good & intermediate & bad & very bad \\
% &  &  &  &  &  \\
\midrule
asphalt & 971 & 1,696 & 821 & (+123) 246  & - \\
concrete & 314 & 350 & 250 &(+4) 58  & - \\
paving stones & 385 & 1,063 & (+322) 519 & (+39) 70 & - \\
sett & - & (+30) 129 & 694 & 540 & - \\
unpaved & - & - & 326 & 387 & 303 \\
\bottomrule
\end{tabular}
\end{table}

\begin{table}[ht]
\caption{Utilized prompt for prompt-based surface quality classification on the example of surface type asphalt.}
\label{tab:prompt}
\begin{tabular}{llp{14cm}}
\toprule
\textbf{role} & \textbf{type} & \textbf{text} \\
\midrule
system & text & You are a data annotation expert trained to classify the   quality level of road surfaces in images. \\
user & text & You need to determine the   quality level of the road surface depicted in the image, following this   defined scale: 
\newline Asphalt surfaces   are graded from excellent to bad according to the following scale:     \newline                    
1) excellent: As   good as new asphalt, on which a skateboard or rollerblades will have no   problem.  \newline              
2) good: Asphalt   showing the first signs of wear, such as narrow, smaller than 1.5 cm cracks,   or wider cracks filled up with tar, shallow dents in which rainwater may   collect, which may cause trouble for rollerblades but not for racing   bikes.\newline
3) intermediate:   Asphalt roads that shows signs of maintenance, such as patches of repaired   surface, wider cracks larger than 2cm. Asphalt sidewalks may contain   potholes, but these are small, shallow (3cm deep) and can be easily   avoided, asphalt driving lanes shows damage due to subsidence (depressions of   a scale \textgreater{}50 cm) or heavy traffic (shallow ruts in asphalt caused by trucks   in summer). This means that the road can be used by normal city bikes,   wheelchairs and sports cars, but not by a racing bike.\newline
4) bad: Damaged   asphalt roads that show clear signs of maintenance: This might include   potholes, some of them quite deep, which might decrease the average speed of   cars.  However, it isn’t so rough that   ground clearance becomes a problem. Meaning that the street causes trouble to   normal city bike but not a trekking bike and a car.
\newline
Please adhere to   the following instructions:\newline
1) Step 1: If you   detect multiple surface types, only consider the path, driving lane, cycleway   or sidewalk in the focus area.\newline
2) Step 2: Check   if the road surface is worn off and if you can find any damages, like   cracks.\newline
3) Step 3: Check   the quantity and the size of the damages.\newline
4) Step 4: Then   decide if you could ride on the surface with    a skateboard, rollerblades, racing bikes, city bike, or a normal   car.\newline
5) Step 5: If you   detect characteristics of two classes, choose the worse class.\newline
How would you   rate this image using one of the four options of the defined scale:\newline
1)   excellent\newline
2) good\newline
3)   intermediate\newline
4) bad\newline
Provide your   rating in one word disregarding the bullet point numbers and brackets as a   string using the four levels of the scale provided.
Make sure you   have the same number of image urls as input as you have output values.
\newline
Do not provide   any additional explanations for your rating; focus solely on the road surface   quality.\\
user & image\_url & {`url':   f`data:image/jpeg;base64,\{excellent\_encoded\_image\}'}\\        
user & text & This was an example for 'excellent' \\
user & image\_url & {`url':   f`data:image/jpeg;base64,\{good\_encoded\_image\}'}\\        
user & text & This was an example for 'good' \\
user & image\_url & {`url':   f`data:image/jpeg;base64,\{intermediate\_encoded\_image\}‘}\\     
user & text & This was an example for 'intermediate' \\
user & image\_url & {`url':   f`data:image/jpeg;base64,\{bad\_encoded\_image\}'}\\             
user & text & This was an example for 'bad' \\
user & text & Please decide now in one word which category the following picture belongs to as instructed in the beginning of the prompt. Then compare this image to the previous ones and decide whether this category is correct. \\
user & image\_url & {`url':   f`data:image/jpeg;base64,\{image\}'}\\   
 \bottomrule
 \end{tabular}
\end{table}

\begin{table}[th]
\centering
\caption{Type model results in terms of precision, recall and F1 scores for StreetSurfaceVis test data. Standard deviations are depicted in parentheses. Combined type results are (unweighted) averages over all classes.  Performance of quality models is measured in terms of accuracy, 1-off accuracy, and Spearman correlation coefficient $\rho$.}
\label{tab:modelEval}
\begin{tabular}{lr|ccc|crc}
\toprule
         & &  \multicolumn{3}{c |}{type model results} &  \multicolumn{3}{c }{quality model results} \\
&support & precision & recall & F1 score  & acc & 1-off acc & $\rho$ \\
\midrule
asphalt       & 530 & .98 \footnotesize{(.00)} & .90 \footnotesize{(.01)} & .94 \footnotesize{(.01)} & .63 \footnotesize{(.02)} & .99 \footnotesize{(.00)}& .58 \footnotesize{(.02)}\\ %MSE & .56 \footnotesize{(.01)}
concrete      & 16 & .23 \footnotesize{(.01)} & .74 \footnotesize{(.07)} & .35 \footnotesize{(.01)} & .65 \footnotesize{(.03)} & 1.00 \footnotesize{(.00)}&  .49 \footnotesize{(.11)}\\ % MSE .51 (.05)
paving stones & 133 & .89 \footnotesize{(.03)} & .92 \footnotesize{(.03)} & .90 \footnotesize{(.02)} & .61 \footnotesize{(.03)} & 1.00 \footnotesize{(.00)}&  .42 \footnotesize{(.03)}\\ % MSE .57 \footnotesize{(.02)}
sett          & 32 & .88 \footnotesize{(.03)} & .92 \footnotesize{(.04)} & .90 \footnotesize{(.03)} & .71 \footnotesize{(.06)} & 1.00 \footnotesize{(.00)}&  .65 \footnotesize{(.06)}\\ % MSE: & .49 (.04)
unpaved       & 85 & .93 \footnotesize{(.02)} & .97 \footnotesize{(.02)} & .95 \footnotesize{(.01)} & .61 \footnotesize{(.05)} & .99 \footnotesize{(.01)}&  .64 \footnotesize{(.04)}\\ % MSE: .57 (.01)&
\hline
combined       & 776 & .78 \footnotesize{(.01)} & .89 \footnotesize{(.02)} & .81 \footnotesize{(.01)} & .63 \footnotesize{(.01)} & .99 \footnotesize{(.00)}&  .72 \footnotesize{(.01)}\\ % MSE:  & .56 (.00)
\bottomrule
\end{tabular}
\end{table}

\begin{table}[ht]
\centering
\caption{Type and quality model results for both setups, where the model is trained on one dataset and predicted on the other. Type results are shown as precision, recall, and F1 scores, combined with (unweighted) averages over all classes. Surface quality results are represented by the Spearman rank correlation coefficient $\rho$. 
Standard deviations are depicted in parentheses. Superior values are indicated in bold and asterisks ($^*$) denote significant differences between models at a significance level of $p < 0.05$.}
\label{tab:crossDatasetRes}
\begin{tabular}{lrcc|rcc}
\toprule
         training&  \multicolumn{3}{c |}{StreetSurfaceVis} & \multicolumn{3}{c}{ RTK}\\
         prediction&  \multicolumn{3}{c |}{RTK} & \multicolumn{3}{c}{StreetSurfaceVis}\\
\hline
         & support & precision | recall | F1 & $\rho$  & support & precision | recall | F1 & $\rho$ \\
\midrule
asphalt     & 3281  &    \textbf{.83\tmark[{\makebox[0pt][l]{*}}]} {\footnotesize(.07)} | .95 {\footnotesize(.05)} | \textbf{.88}\tmark[{\makebox[0pt][l]{*}}] {\footnotesize(.04)}& \textbf{.68\tmark[{\makebox[0pt][l]{*}}] } {\footnotesize(.02)}  &
               4160 &    .70\tmark[{\makebox[0pt][l]{*}}] {\footnotesize(.06)} | \textbf{.98} {\footnotesize(.02)} | .81\tmark[{\makebox[0pt][l]{*}}] {\footnotesize(.03)}&  \enspace{.25\tmark[{\makebox[0pt][l]{*}}]  {\footnotesize(.06)}}  \\
paved       & 1627  &    \textbf{.94} {\footnotesize(.07)} | .60 {\footnotesize(.12)} | .72 {\footnotesize(.08)}& \textbf{.43\tmark[{\makebox[0pt][l]{*}}] } {\footnotesize(.03)}   &                       3255 &  .89 {\footnotesize(.07)} | \textbf{.65} {\footnotesize(.13)} | \textbf{.74} {\footnotesize(.09)} & -.04\tmark[{\makebox[0pt][l]{*}}] {\footnotesize(.07)}   \\
unpaved     & 1389  & \textbf{.85} {\footnotesize(.13)} | \textbf{.86}\tmark[{\makebox[0pt][l]{*}}] {\footnotesize(.17)} | \textbf{.84}\tmark[{\makebox[0pt][l]{*}}] {\footnotesize(.06)}  & \textbf{.32\tmark[{\makebox[0pt][l]{*}}] } {\footnotesize(.14)}   & 
                931 & .78 {\footnotesize(.09)} | .06\tmark[{\makebox[0pt][l]{*}}] {\footnotesize(.04)} | .11\tmark[{\makebox[0pt][l]{*}}] {\footnotesize(.08)}  &  -.15\tmark[{\makebox[0pt][l]{*}}] {\footnotesize(.04)}  \\
\hline
\Tstrut
combined     & 6297 & \textbf{.87\tmark[{\makebox[0pt][l]{*}}] }{\footnotesize(.02)} | \textbf{.80\tmark[{\makebox[0pt][l]{*}}] }{\footnotesize(.07)} | \textbf{.81\tmark[{\makebox[0pt][l]{*}}] }{\footnotesize(.06)} & \textbf{.52\tmark[{\makebox[0pt][l]{*}}] } {\footnotesize(.09)} & 
8346 & .79\tmark[{\makebox[0pt][l]{*}}] {\footnotesize(.03)} | .56\tmark[{\makebox[0pt][l]{*}}] {\footnotesize(.05)} | .56\tmark[{\makebox[0pt][l]{*}}] {\footnotesize(.05)} &  \enspace{.16\tmark[{\makebox[0pt][l]{*}}] {\footnotesize(.11)}}\\
\bottomrule
\end{tabular}
\end{table}

\end{document}